\documentclass[10pt,twocolumn,letterpaper]{article}

\usepackage{iccv}
\usepackage{times}
\usepackage{epsfig}
\usepackage{graphicx}
\usepackage{amsmath}
\usepackage{amssymb}

\usepackage{caption}
\usepackage{enumitem}
\usepackage{bm}

\usepackage{booktabs}
\usepackage[dvipsnames]{xcolor} 
\usepackage{multirow}
\usepackage{colortbl}
\usepackage{pifont}

\definecolor{mygray}{gray}{.9}
\definecolor{mygray2}{gray}{.88}
\definecolor{mygray3}{gray}{.85}

\definecolor{citecolor}{RGB}{34, 149, 34}
\usepackage[pagebackref=true,breaklinks=true,letterpaper=true,colorlinks,citecolor=citecolor,bookmarks=false]{hyperref}

\newcommand{\tablestyle}[2]{\setlength{\tabcolsep}{#1}\renewcommand{\arraystretch}{#2}\centering\footnotesize}

\newcommand{\xmark}{\ding{55}}%

\iccvfinalcopy 

\def\iccvPaperID{4222} 

\ificcvfinal\fi

\begin{document}
\title{{\texttt{FerKD}}: Surgical Label Adaptation for Efficient Distillation}

\author{Zhiqiang Shen\\
Mohamed bin Zayed University of AI\\
{\tt\small Zhiqiang.Shen@mbzuai.ac.ae}
}

\maketitle
\ificcvfinal\thispagestyle{empty}\fi

\begin{abstract}
We present \texttt{FerKD}, a novel efficient knowledge distillation framework that incorporates partial soft-hard label adaptation coupled with a region-calibration mechanism. Our approach stems from the observation and intuition that standard data augmentations, such as RandomResizedCrop, tend to transform inputs into diverse conditions: easy positives, hard positives, or hard negatives. In traditional distillation frameworks, these transformed samples are utilized equally through their predictive probabilities derived from pretrained teacher models. However, merely relying on prediction values from a pretrained teacher, a common practice in prior studies, neglects the reliability of these soft label predictions. To address this, we propose a new scheme that calibrates the less-confident regions to be the context using softened hard groundtruth labels. Our approach involves the processes of {\bf \em hard regions mining + calibration}. We demonstrate empirically that this method can dramatically improve the convergence speed and final accuracy. Additionally, we find that a consistent mixing strategy can stabilize the distributions of soft supervision, taking advantage of the soft labels. As a result, we introduce a stabilized {SelfMix} augmentation that weakens the variation of the mixed images and corresponding soft labels through mixing similar regions within the same image. \texttt{FerKD} is an intuitive and well-designed learning system that eliminates several heuristics and hyperparameters in former FKD solution~\cite{shen2022fast}. More importantly, it achieves remarkable improvement on ImageNet-1K and downstream tasks. For instance, \texttt{FerKD} achieves 81.2\% on ImageNet-1K with ResNet-50, outperforming FKD and FunMatch by remarkable margins. Leveraging better pre-trained weights and larger architectures, our finetuned ViT-G14 even achieves 89.9\%. Our code is available at \url{https://github.com/szq0214/FKD/tree/main/FerKD}.
\end{abstract}

\begin{figure}
\begin{center}
\includegraphics[width=0.481\textwidth]{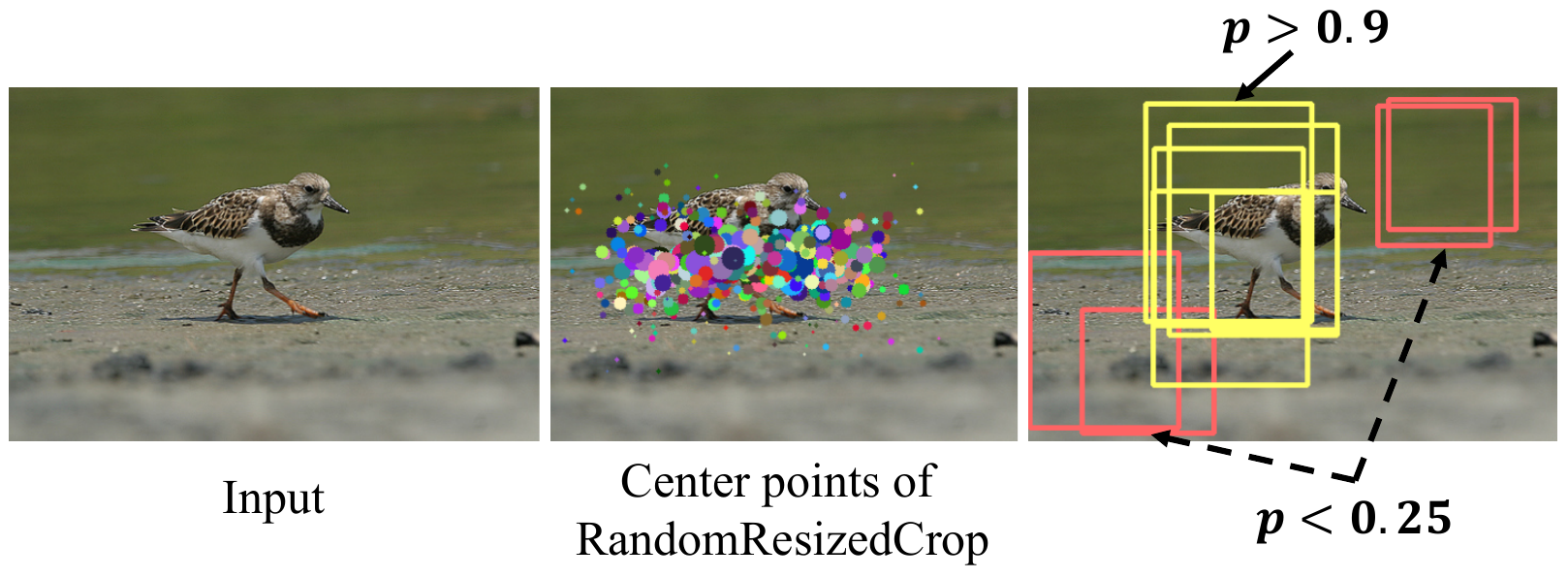}
\end{center}
\vspace{-0.2in}
   \caption{Illustration of motivation for \texttt{FerKD}. The left figure depicts the original input, and the middle figure shows the center points of bounding boxes generated using {\texttt{RandomResizedCrop}}. The radius of each circle corresponds to the area of the bounding box. It can be observed that the center points of the bounding boxes are concentrated in the center of the image, and their area increases as they approach the center. The right figure displays several top and bottom confident bounding boxes and their corresponding predictive probabilities from a pre-trained teacher or teachers ensemble. The proposed hard region calibration strategy is established based on these predictions.}
\label{fig:short}
\end{figure}

\section{Introduction}

Knowledge Distillation (KD)~\cite{hinton2015distilling} has achieved impressive results in various visual domains, including image classification~\cite{xie2020self,shen2020meal,beyer2021knowledge,shen2022fast}, object detection~\cite{chen2017learning,wang2019distilling,guo2021distilling,dai2021general,yang2022focal} and semantic segmentation~\cite{liu2019structured,hou2022point,Ji_2022_CVPR}. However, KD methods are often computationally expensive and inefficient due to the additional computational burden imposed by the teacher models. The primary advantage of KD that motivates its usage is its ability to generate precise soft labels that convey more informative details about the input examples. It differs from other label softening techniques, such as label smoothing~\cite{szegedy2016rethinking}, Mixup~\cite{zhang2018mixup}, and CutMix~\cite{yun2019cutmix}, mainly in two aspects: (1) KD generates soft labels dynamically in each iteration, which is more informative than fixed smoothing patterns used in label smoothing; (2) Mixup and CutMix techniques essentially combine hard labels with coefficients, while KD produces soft labels that are highly correlated with the input sample. This allows KD's soft labels to become more accurate when different data augmentations, such as {\em RandomResizedCrop}, {\em flipping and rotation}, {\em color jittering}, etc., are applied. In general, mixing-based label softening methods cannot monitor such changes in input content, but KD can address them effortlessly.

To overcome the computational inefficiency of traditional knowledge distillation, FKD~\cite{shen2022fast} was developed to generate region-level soft labels in advance and reuse them across multiple training cycles to eliminate redundant computation. This approach only requires the preparation of soft labels once at the beginning and they can be reused indefinitely. However, this approach overlooks certain critical issues. One is the quality of the soft labels. When using {\em RandomResizedCrop} to generate regions, some may be cropped from background areas, and the teacher model will still produce a soft label for them based on their similarity to the dataset classes. However, in some cases, these areas may contain irrelevant noise, compensatory information, or context information for the class, and the soft labels may not accurately reflect the context of information they carry. To address this problem, this work proposes to recalibrate these soft labels by incorporating context information from hard ground-truth labels with smoothing.

\begin{figure}[t]
\begin{center}
\includegraphics[width=0.48\textwidth]{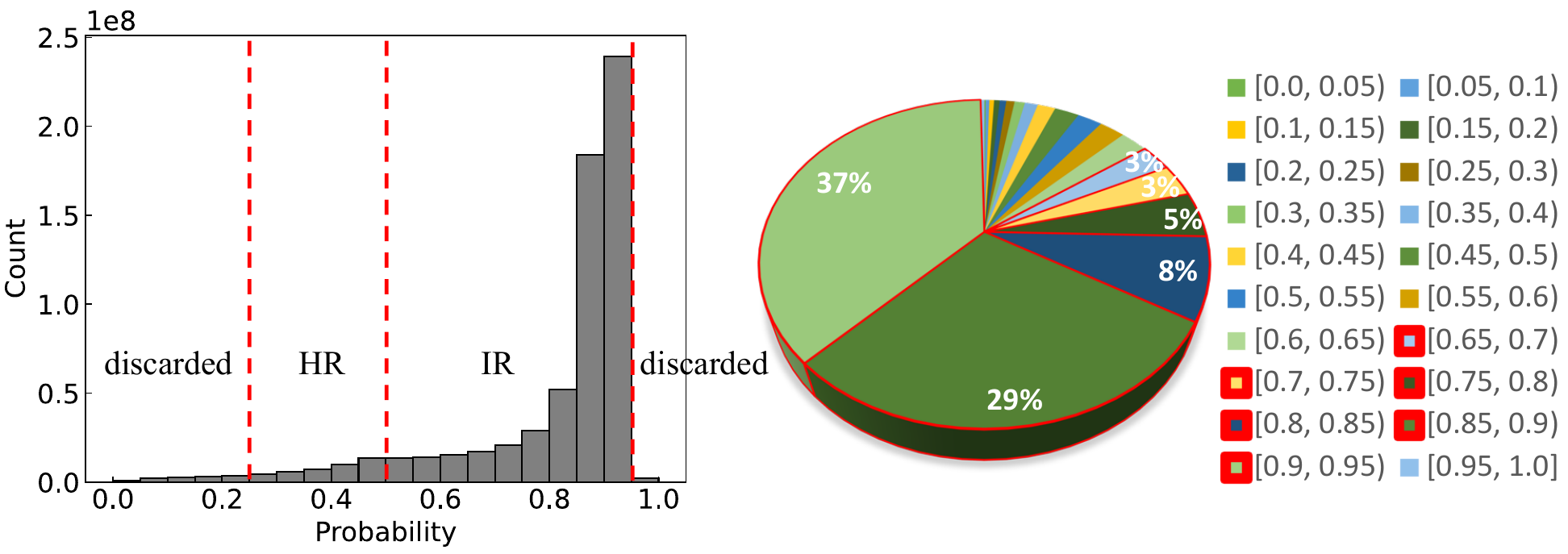}
\end{center}
\vspace{-0.25in}
   \caption{Statistics of soft label max-probability for crops on ImageNet-1K. The soft label is from {FKD}~\cite{shen2022fast}. In each image, 500 regions are randomly cropped.}
\label{fig:Statistics}
\vspace{-0.2in}
\end{figure}

Furthermore, due to the random nature of the sampling process, a certain proportion of crops that are either excessively easy or difficult do not contribute to the model's learning capacity. As demonstrated in Fig.~\ref{fig:Statistics} and Table~\ref{tab:statistics}, these samples can be discarded to expedite the convergence process. The pre-generated soft labels can be utilized as useful indicators to select these specific samples. In our adaptation of surgical soft labeling, we categorize the soft labels into four distinct groups: extreme hard (negative), moderate hard (background or context), hard positive (partial object), and easy positive. Each of these categories is subject to different treatment methodologies.

\noindent{\textbf{The Role of Background.}} The role of the background in images is essential, as it provides critical context and spatial information that aids the model in accurately identifying objects of interest within the scene. Backgrounds can vary in complexity and structure, ranging from simple monochromatic backgrounds to highly cluttered and detailed ones. Soft labels in background areas are typically low, and therefore, it is crucial to handle the background carefully with precise supervision to achieve higher model capability within our surgical label calibration framework.

\noindent{\textbf{Hard Regions Mining and Calibration.}} Hard Regions Mining involves the identification and isolation of challenging or complex regions within an image that the model struggles to identify accurately. These regions can include objects with complex shapes, occlusions, or those with low contrast. By identifying these regions, the model can focus on learning the features and characteristics of these regions, resulting in improved performance. Calibration, on the other hand, involves adjusting the confidence levels of the model's predictions in challenging regions. The model's predictions may be less reliable in hard regions, leading to lower confidence scores. Calibrating the predictions can improve the model's accuracy in these regions by adjusting the confidence levels of the predictions. We found that carefully discarding a portion of negative crops and selecting those hard positive crops by calibrating their labels, can force the training process more efficient and effective.

\noindent{\textbf{Stable Training on Soft Labels.}} Mixture-based augmentations, such as Mixup and CutMix have seen widespread use for training models under hard supervision, where each image is labeled with a single class label. However, in the soft label scenario, we have made a different observation: when employed together with pre-generated soft labels on a typical ResNet, Mixup and CutMix tend to be overly strong, which, conversely, leads to decreased performance. To mitigate label fluctuations and achieve more stable training, we propose a \texttt{SelfMix} scheme, which is particularly suitable for cases where data augmentation should not be so strong, such as in finetuning distillation, where mixture-based augmentation is usually disabled. On the other hand, when training ViT models from scratch, stronger data augmentation can yield better results~\cite{touvron2021training,steiner2021train}, which is consistent with the larger capacity perspective of this type of network.

\begin{table}[t]
\centering
\resizebox{0.38\textwidth}{!}{
\begin{tabular}{l|l|l|l}
range ($P$)      & ratio           & range ($P$)  & agg. ratio  \\ \midrule[1pt]
$[0.0, 0.1)$    &    0.43\%   &  $[0, 0.1)$     &  0.43\%  \\  
 $[0.1, 0.2)$    &    0.89\%   &  $[0, 0.2)$     &  1.32\%  \\ 
$[0.2, 0.3)$    &     1.29\%   &  $[0, 0.3)$     &  2.61\% \\  
 $[0.3, 0.4)$    &    2.03\%   &  $[0, 0.4)$     &  4.64\%  \\ 
 $[0.4, 0.5)$    &    3.66\%   &  $[0, 0.5)$     &  8.31\%  \\  
 $[0.5, 0.6)$    &    4.35\%   &  $[0, 0.6)$     &  12.65\%  \\ 
 $[0.6, 0.7)$    &    5.04\%   &  $[0, 0.7)$     &  17.69\%  \\  
 $[0.7, 0.8)$    &    7.76\%   &  $[0, 0.8)$     &  25.45\%  \\ 
 $[0.8, 0.85)$    &   8.14\%   &  $[0, 0.85)$    &  33.59\%  \\  
$[0.85, 0.9)$    &    28.73\%   &  $[0, 0.9)$    &  62.32\%  \\ 
 $[0.9, 0.95)$    &   37.34\%   &  $[0, 0.95)$   &  99.67\%  \\ 
 $[0.95, 1.0)$    &   0.33\%   &  $[0, 1.0)$     &  100\%  \\  
\end{tabular}
}
\vspace{-0.1in}
\caption{Detailed statistics of soft labels. ``range'' indicates max-probability of crops, ``ratio'' indicates the percentage in the whole crops. ``agg. ratio'' is the aggregated ratio.}
\label{tab:statistics}
\vspace{-0.15in}
\end{table}

In summary, our contributions of this work are:

\noindent \ \ - We present \texttt{FerKD}, a sample-calibration framework for {\em Faster Knowledge Distillation} that achieves state-of-the-art performance. We conduct extensive analysis, ablation, and discussion on the impact of hard and easy samples.

\noindent \ \ - We make two key observations in the pre-generated soft label training framework. Firstly, we observe that the few most challenging and simplest crops obtained through the \texttt{ RandomResizedCrop} operation do not contribute significantly to the model's learning and can therefore be removed. Secondly, we find that moderately hard crops can provide crucial contextual information that improves the model's ability to learn robust representations.

\noindent \ \ - We perform extensive experiments on ImageNet-1K and downstream tasks. On ImageNet-1K, \texttt{FerKD} achieves an accuracy of 81.2\% using the ResNet-50. When leveraging self-supervised pre-trained weights, our larger model fine-tuned using ViT-G/14 achieves an accuracy of 89.9\%.

\section{Related Work}

\noindent{\textbf{Knowledge Distillation and Fast Knowledge Distillation.}} Knowledge Distillation~\cite{hinton2015distilling} is a learning method in which a ``student'' model is trained to imitate the predictions of a larger, more complex ``teacher'' model. A key advantage of this approach is that the teacher model can provide soft supervision that contains more information regarding the input data than traditional one-hot human annotated labels, particularly when the input data is subject to data augmentation. There have been many recent variants and extensions of knowledge distillation~\cite{romero2014fitnets,papernot2016distillation,wang2018dataset,zhang2019your,muller2019does,park2019relational,shen2021label,chung2020feature,stanton2021does,wu2022tinyvit,touvron2021training,li2022self,xiaolong2023norm}, including approaches that use internal feature representations~\cite{chung2020feature}, adversarial training with discriminators~\cite{shen2019meal}, transfer learning techniques~\cite{yim2017gift}, fast distillation~\cite{shen2022fast} via instance label preparation, and methods that prioritize patient and consistent learning~\cite{beyer2021knowledge}. 

\noindent{\textbf{Hard Sample Mining.}} The aim of hard sample mining~\cite{loshchilov2015online,shrivastava2016training} is to enhance the performance of learning models by selectively focusing on challenging examples that are typically difficult to classify. By prioritizing hard samples during training, models can be trained to better handle a wider range of real-world scenarios and improve their overall performance. One approach~\cite{shrivastava2016training} to achieve this for object detection is to use Online Hard Example Mining (OHEM) which employs a strategy of selecting challenging examples during training of region-based ConvNet detectors. The motivation behind this approach is that detection datasets typically comprise an overwhelming number of easy examples and a small number of hard examples. Automatic selection of these hard examples can make training more effective and efficient.

Some other techniques that are close to hard sample mining: (1) Curriculum learning~\cite{bengio2009curriculum,wang2021survey,soviany2022curriculum}: it trains a model on easy examples first and then gradually increasing the difficulty of the examples over time. (2) Active learning~\cite{settles2009active,michael2006s,prince2004does}: it selects the most informative or uncertain samples for labeling by a human annotator. By focusing on the samples that the model is most uncertain about, the model can learn to better generalize and improve its performance on difficult samples. (3) Loss functions: Attentive loss functions can be used to emphasize the importance of hard samples during training. For example, focal loss~\cite{lin2017focal} places more weight on the difficult examples during training, helping the model to learn to handle them better. 

\noindent{\textbf{Data Augmentations.}} Several studies have incorporated data augmentations into distillation frameworks to improve performance. For instance, FunMatch~\cite{beyer2021knowledge} employed Mixup, and FKD~\cite{shen2022fast} utilized CutMix. Both techniques achieved competitive accuracy on large-scale ImageNet-1K dataset. In this work, we investigate the impact of data augmentation intensity on soft labels. We discover that different network architectures require unique data augmentation levels. Specifically, ResNet necessitates mild data augmentation, while ViTs require stronger data augmentation. However, even for ViTs, finetuning distillation requires a reduction in the intensity of data augmentation, especially for mixture-based methods. Motivated by this observation, we propose a mild \texttt{SelfMix} approach for ResNet and finetuning distillation scenarios.

\section{Approach}
The proposed {\texttt{FerKD}} is a soft label calibration framework for fast and efficient knowledge distillation training. In this section, we aim to provide an elaborated overview of our method, starting with an in-depth analysis of the roles of hard and soft labels in the distillation process. We then present the key components of our approach, which include a region selection strategy and a soft label calibration scheme. Additionally, we explore the data augmentation requirements for soft labels and introduce a simple label ensemble technique to enhance the quality of the soft labels.

\subsection{Revisiting Hard and Soft Label in Distillation}

The utilization of hard and soft labels is dependent on the comprehension, problem scenarios, and underlying objectives. Various arguments exist regarding their effectiveness. The vanilla knowledge distillation method~\cite{hinton2015distilling} employs both hard and soft labels to maximize the benefits of both. However, recent researches~\cite{shen2020meal,shen2021label,yun2021re,shen2022fast} suggest that the use of hard labels is not necessary in distillation on large-scale datasets as strong teachers can provide more precise soft supervision. Incorporating hard labels may introduce erroneous supervisory signals, ultimately hampering student performance. Other than the above views, this work presents a novel perspective beyond them by acknowledging that both hard and soft labels offer unique advantages, highlights the proper practice of usage that is necessary, and finally proposes an elegant solution to reap benefits derived from both sources of hard and soft labels.

\noindent{\textbf{Combination of Soft and Hard Labels.}}  
In vanilla KD design~\cite{hinton2015distilling}, for each training example, it will minimize two loss terms for both hard and soft labels. The final objective $\mathcal L_\text{\em VKD}$ can be formulated as:
\begin{equation}
\small
\begin{aligned}
\mathcal L_\text{\em VKD}\! =\! \frac{1}{N}\sum_{ x} (\alpha\!*\!\underbrace{\mathcal L_{h}(p_\theta(x),y_h(x))}_{\textbf {CE loss with hard label}}\! + (1\!-\!\alpha)\!*\!\underbrace{\mathcal L_{s}(p_\theta(x),y_s(x))}_{\textbf {KL loss with soft label}})
 \end{aligned}
\end{equation}
where $\alpha$ is the coefficient to balance the loss signal intensity from soft and hard labels. $p_\theta(x)$ is the logits prediction from the student model and $\theta$ is its parameters. $y_h$ is the hard label and $y_s$ is the soft label from a pre-trained teacher. $N$ is the total number of training samples $x$.

\noindent{\textbf{Full Soft Label Training.}} 
Fast KD~\cite{shen2022fast} proposes to use the soft label solely since the prediction from strong teachers is precise enough. Thus, the loss objective $\mathcal L_\text{\em FKD}$ is:
\begin{equation}
\mathcal L_\text{\em FKD} = \frac{1}{N}\sum_x \underbrace{\hat {\mathcal L}_{s}(p_\theta(x),y_s(x))}_{\textbf {SCE loss with soft label}}
\end{equation}
where ``SCE'' is the soft version of cross-entropy loss.

\begin{figure}[t]
\begin{center}
\includegraphics[width=0.42\textwidth]{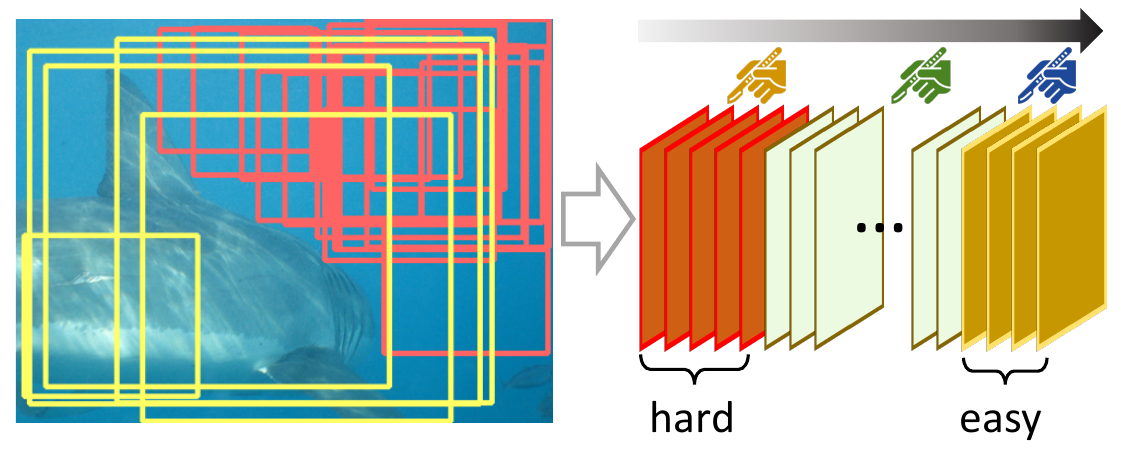}
\end{center}
\vspace{-0.2in}
   \caption{Illustration of region calibration according to their predictive probabilities in \texttt{FerKD}. Left is the input image with {\texttt{RandomResizedCrop}}. \textcolor{Goldenrod}{Bounding box} is with high probability and \textcolor{red}{bounding box} is with low probability.} 
\label{fig:calboration}
\vspace{-0.15in}
\end{figure}

\subsection{{\bf\texttt{FerKD}}: Surgical Label Calibration Distillation}
\begin{figure}[h]
\begin{center}
\vspace{-0.15in}
\includegraphics[width=0.48\textwidth]{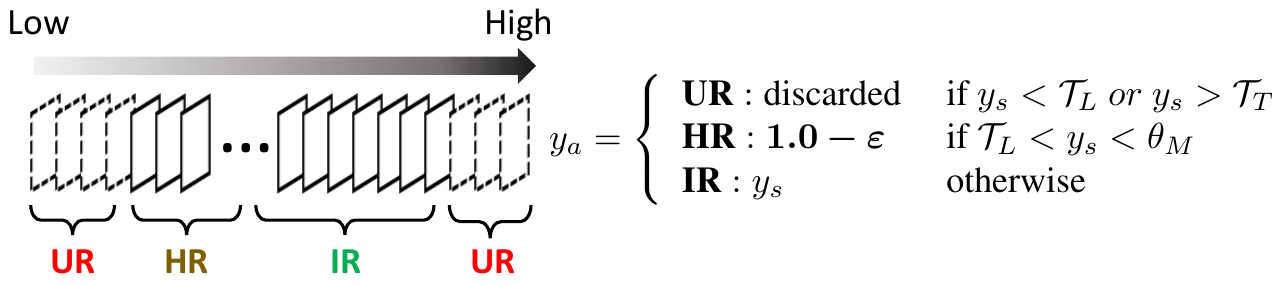}
\end{center}
\vspace{-0.2in}
   \caption{Illustration of region calibration according to their predictive probabilities in \texttt{FerKD}. Left is the input image with {\texttt{RandomResizedCrop}}. Right is the rule for calibrating the probabilities of regions.} 
\label{fig:selection}
\vspace{-0.2in}
\end{figure}

\begin{figure}[h]
\begin{center}
\vspace{-0.1in}
\includegraphics[width=0.45\textwidth]{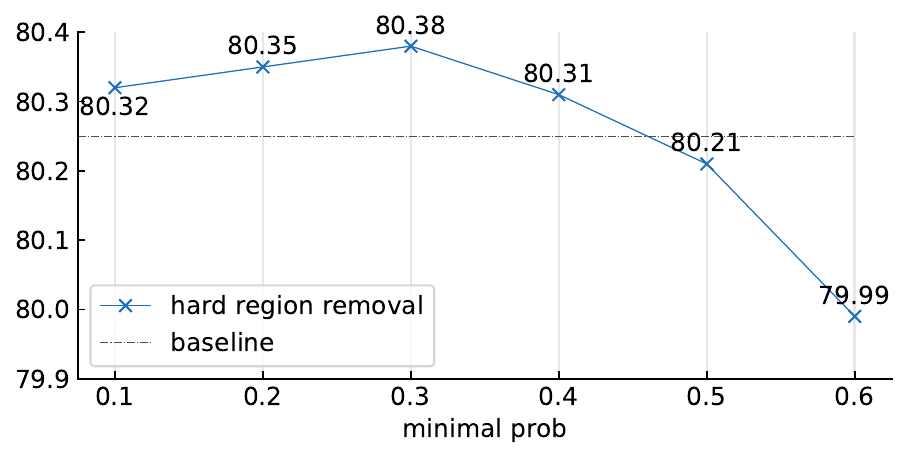}
\includegraphics[width=0.45\textwidth]{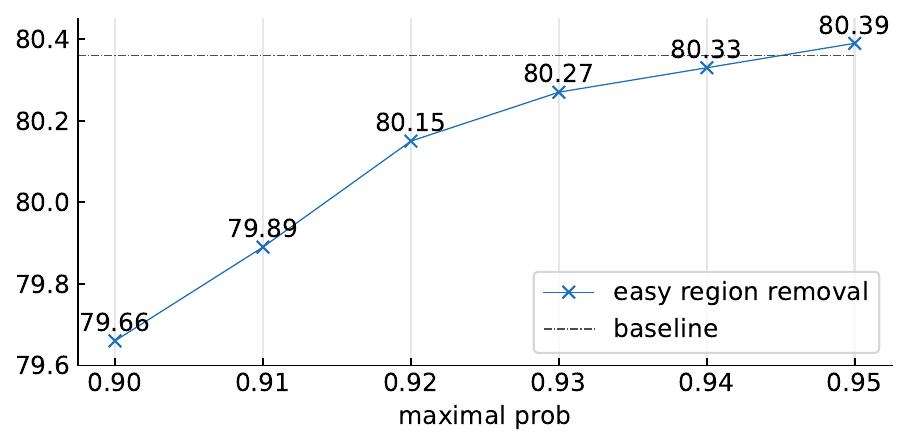}
\end{center}
\vspace{-0.25in}
   \caption{{\bf Minimal and maximal probability}. The upper figure indicates that only regions having the max probability in $[\text{minimal}, 1.0]$ will be trained, and {\em baseline} indicates that the model is trained with all randomly sampled regions. The bottom figure indicates that only regions having the max probability in $[0.3, \text{maximal}]$ will be trained, and {\em baseline} indicates that the model is trained with regions in $[0.3, 1.0]$.} 
\label{fig:quick_exp}
\vspace{-0.1in}
\end{figure}

\noindent{\textbf{Surgical/Partial Soft and Hard Label Adaptive Training.}} Different from VKD, the proposed \texttt{FerKD} strategy will only involve one loss term but also unlike FKD, \texttt{FerKD} will employ both hard and soft labels in a single objective term and exploit the additional information derived from both sources. The solution is that we only keep their original soft labels for positive regions since they contain fine-grained information regarding the crops, for those background or context regions, we will re-calibrate them by the human-annotated ground-truth labels to avoid misinformation from the soft labels. Hence, the loss function will be:
\begin{equation}
\mathcal L_\text{\em FerKD} = \frac{1}{N}\sum_x \underbrace{\mathcal L_{{\texttt{adap}:\ \text{h} \ \text{or}\ \text{s}?}}(p_\theta(x),y_a(x))}_{\textbf {SCE loss with hard/soft label}}
\end{equation}
where ``\texttt{adap}:'' indicates$\ \{{hard}\} \ \text{or}\ \{{soft}\}$ used for individual training samples. $y_a$ is the calibrated soft labels. As illustrated in Fig.~\ref{fig:calboration}, we will calibrate regions' soft labels $y_a$ using the following rule: 
\begin{equation}
 y_{a}=\left\{\begin{array}{ll} \textbf{UR}: \text{discarded} & \text { if }  y_s<\mathcal{T}_{L}\ or \   y_s > \mathcal{T}_{T} \\ \textbf{HR}: \bm {1.0-\varepsilon} & \text { if } \mathcal{T}_{L}< y_s <\theta_{M} \\ \textbf{IR}:  y_s & \text { otherwise }\end{array}\right.
\end{equation}
where $\mathcal{T}_L$, $\mathcal{T}_M$, and $\mathcal{T}_T$ are thresholds at low, middle, and top boundaries. ``\textbf{UR}'' represents the uninformative regions such as black or white blocks in an image that will be discarded during training. As shown in Fig.~\ref{fig:selection}, ``\textbf{HR}'' represents the hard regions with a smoothing value $\bm \varepsilon$ and ``\textbf{IR}'' represents the important regions. 
Thus, the key goal in \texttt{FerKD} becomes to identify the positive or background regions.
Thanks to FKD~\cite{shen2022fast}, we can have access to all crops' individual predictions. A quick exploration is performed for verification and the result is shown in Fig.~\ref{fig:quick_exp}. It is clear that by discarding a certain ratio of samples (hardest and easiest), the performance is consistently improved.

\subsection{Training Speed of \bf \texttt{FerKD}}

When training a model, easy examples allow the model to quickly learn the patterns in the data and update its parameters in a way that minimizes the loss function. This means that the model will converge faster and require fewer iterations to reach a satisfactory level of accuracy. On the other hand, hard examples can slow down the convergence speed but they can force models to learn more robust classify boundaries. 
The speed of model convergence can be impacted significantly by the sampling strategy of hard and easy examples. In our \texttt{FerKD} framework, ``hard'' and ``easy'' examples refer to the level of difficulty a particular sample presents to a model. This difficulty can be quantified by the probability assigned to the correct label by the teacher. An ``easy'' sample is one where the probability assigned to the correct label is high, indicating that the teacher is confident in its prediction. On the other hand, a ``hard'' sample is one where the probability assigned to the correct label is low, indicating that the teacher is uncertain about its prediction. The threshold for what constitutes an ``easy'' or ``hard'' sample can vary depending on the specific task and model being used, which is the key for exploring in \texttt{FerKD}.

\begin{figure}[h]
\begin{center}
\vspace{-0.17in}
\includegraphics[width=0.33\textwidth]{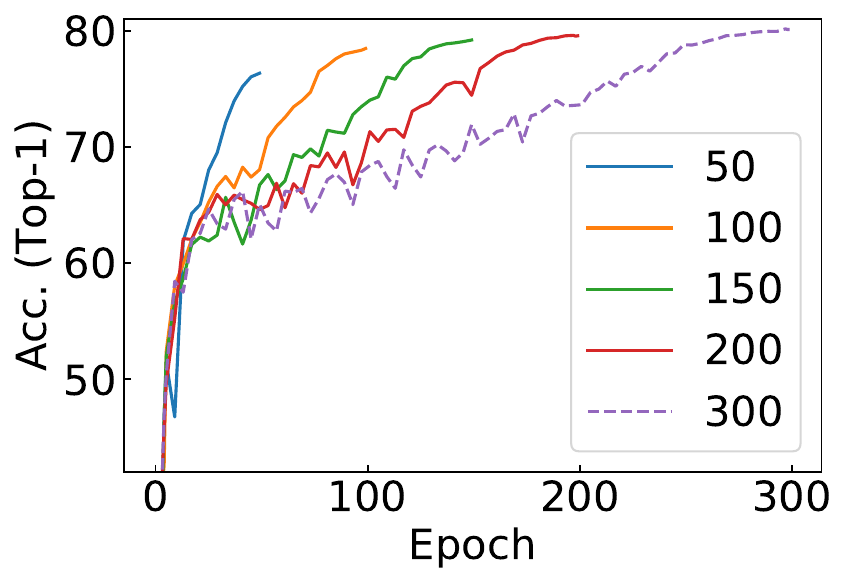}
\end{center}
\vspace{-0.3in}
   \caption{Illustration of the testing accuracy curves. ``300'' represents the training with the full budget. ``50, 100, 150, 200'' are the training with reduced budgets.}
\label{fig:fast_convergence}
\vspace{-0.15in}
\end{figure}

Our sampling strategy involves the removal of the easiest and hardest training examples to reduce computational costs incurred during uninformative training steps. This strategy offers a clear benefit, as shown in Fig.~\ref{fig:fast_convergence}, which presents the test accuracy curves for different training budgets. The results show that, with a training budget reduced by $\frac{2}{3}$ (200 epochs), the achieved accuracy is comparable to that of full-budget training. Although further reducing the budget slightly affects accuracy, our proposed \texttt{FerKD} method is shown to be robust across different training budgets. Moreover, our calibration process will not involve additional training cost since it can be done offline in advance.

\subsection{\bf\texttt{SelfMix}: A Mild and Stable Data Augmentation for Soft Labels}

\begin{figure}[t]
\begin{center}
\includegraphics[width=0.43\textwidth]{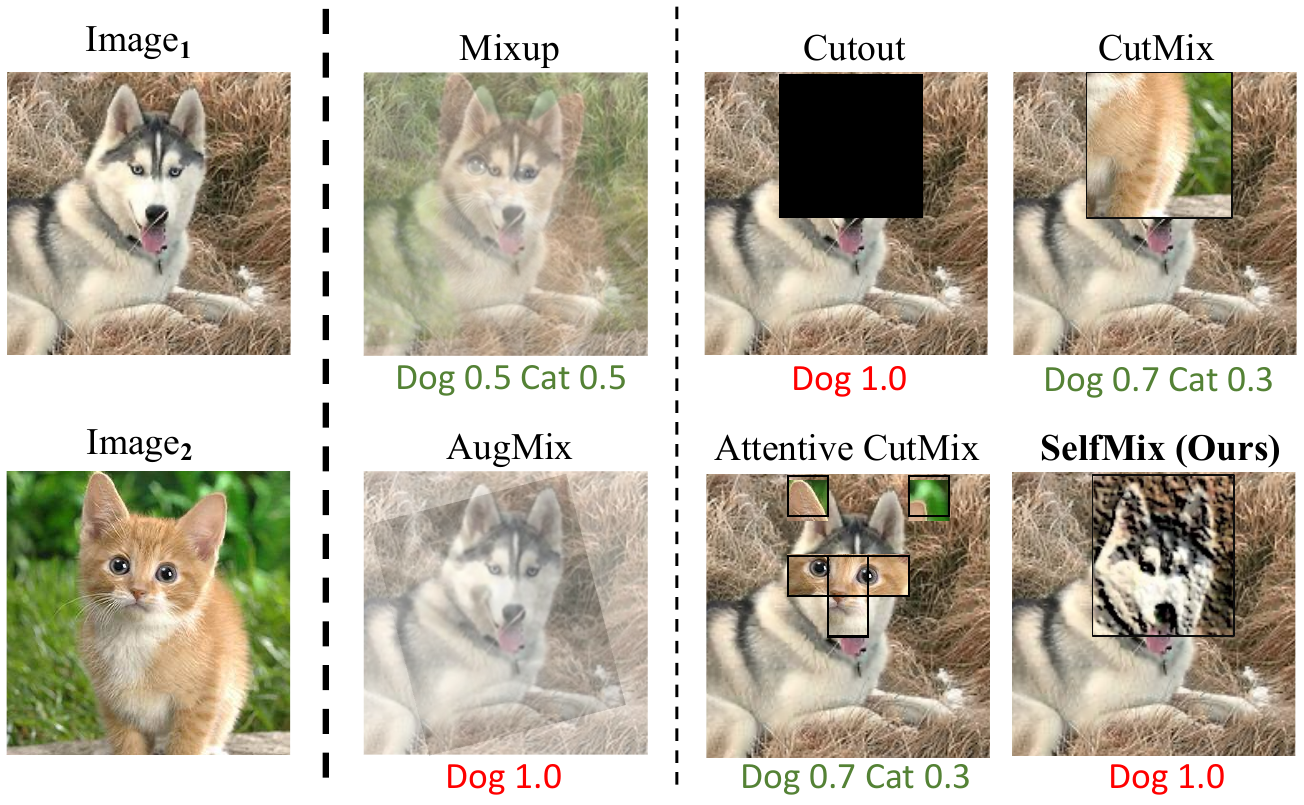}
\end{center}
\vspace{-0.25in}
   \caption{Illustration of the high-level outline for \texttt{SelfMix} to make mixed soft label consistent, reducing the variance. Practically, each label will be a soft distribution for the input instead of the hard label as illustrated, thus it is more moderate on supervision in training. Empirically, it is observed that this strategy is crucial for ConvNet like ResNet but ViT needs intense data augmentations as introduced in~\cite{touvron2021training}.}
\label{fig:motivation_selfmix}
\vspace{-0.1in}
\end{figure}

\noindent{\textbf{Soft label Calibration with \texttt{SelfMix}.}} The current prevailing data augmentation techniques are designed for hard labels or smoothed hard labels. However, in the dynamic soft label scenario, a different approach is necessary to meet the unique attributes of pre-generated soft labels. Soft labels themselves can mitigate overfitting, making it imperative to develop tailored data augmentation techniques that account for soft labels to achieve improved accuracy. The proposed \texttt{SelfMix} is based on the empirical observations presented in Table~\ref{tab:selfmix_result}, which indicate that strong data augmentation does not necessarily improve accuracy if the network is already saturated and may even hurt performance. Consequently, we aim to reduce the intensity of data augmentation while still benefiting from its effects. To achieve this goal, we redesign the data augmentation using a self-mixing approach, as illustrated in Fig.~\ref{fig:motivation_selfmix}.

\begin{figure}[h]
\begin{center}
\vspace{-0.15in}
\includegraphics[width=0.45\textwidth]{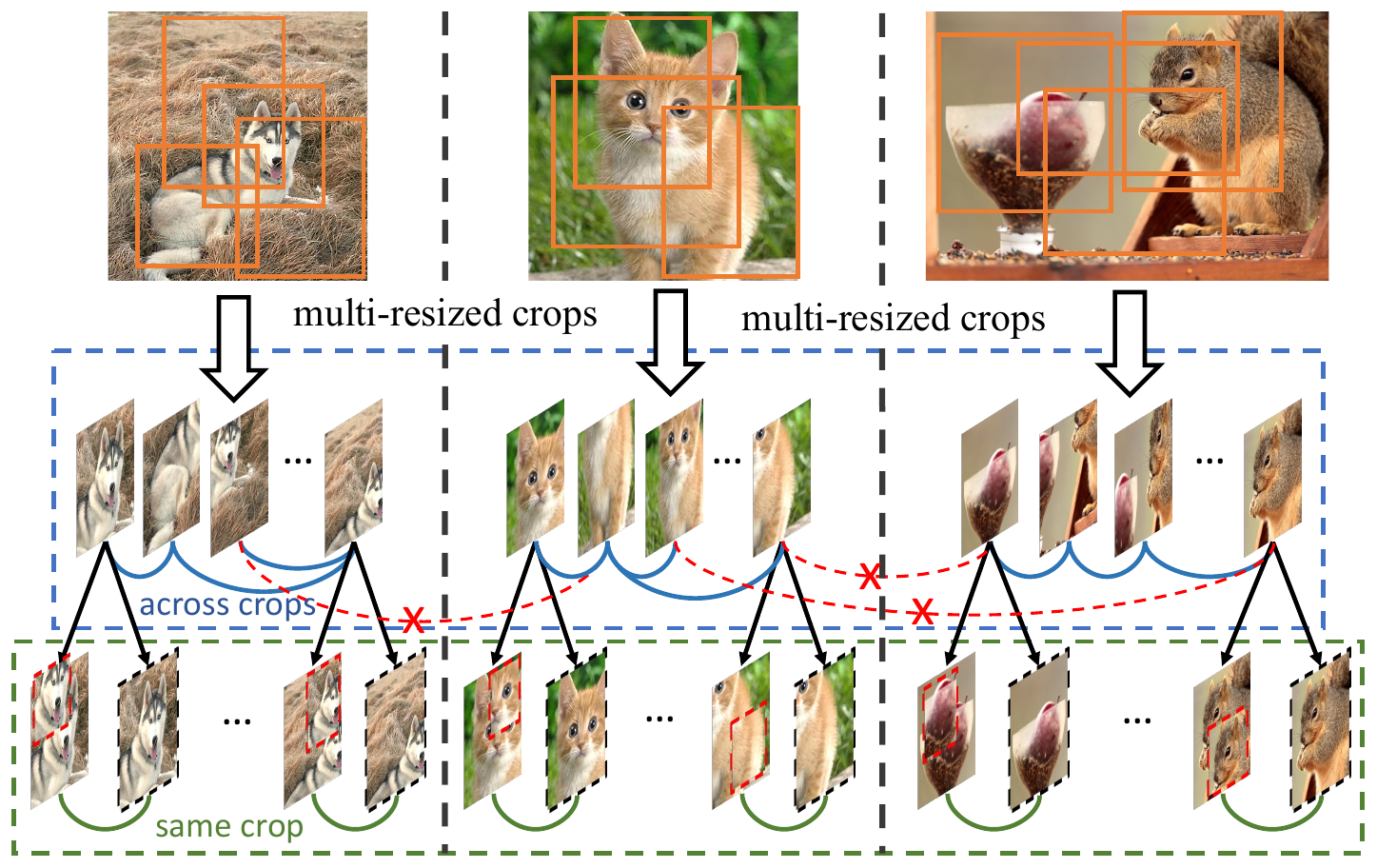}
\end{center}
\vspace{-0.23in}
   \caption{Illustration of the detailed \texttt{SelfMix} augmentation for \texttt{FerKD}. In this strategy, mixture operation only happens within the same image in a mini-batch and cross-image mixing is disabled to preserve stability of soft labels.}
\label{fig:detail_selfmix}
\vspace{-0.13in}
\end{figure}

The specific steps involved in the \texttt{SelfMix} process are illustrated in Fig.~\ref{fig:detail_selfmix}. Mixing operations are exclusively conducted within each individual image to minimize variations between mixed images and their corresponding mixed soft labels. This straightforward constraint yields significant improvements in the performance of the ResNet backbone and in finetuning distillation.

\begin{figure*}[t]
\begin{center}
\includegraphics[width=0.96\textwidth]{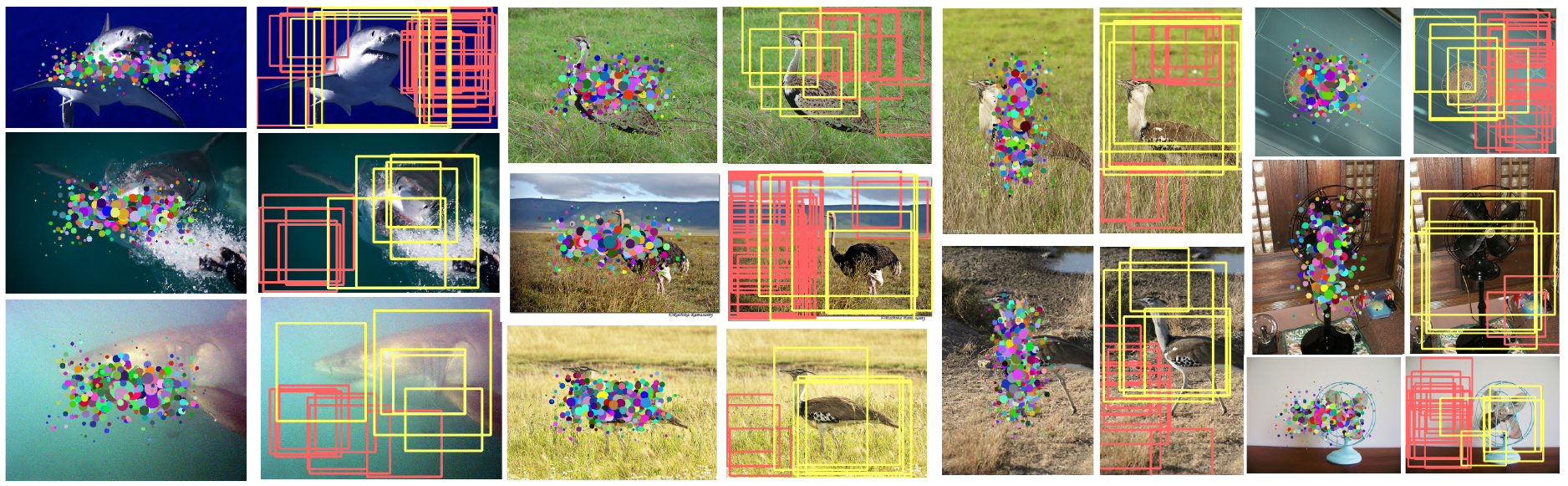}
\end{center}
\vspace{-0.23in}
   \caption{Illustration of the location points from {\em RandomResizedCrop} and the identified crops by teacher model for \textcolor{red}{hard} and \textcolor{yellow}{easy} samples. Note that we do not involve any localization information, but teacher's probability can reflect it automatically.}
\label{fig:more_vis}
\vspace{-0.2in}
\end{figure*}

\noindent{\textbf{Ensemble Supervisions.} It has been demonstrated that ensembling more teachers~\cite{hinton2015distilling,shen2019meal} can enhance the performance of distilled students. In this work, we apply this approach to improve the quality of pre-existing soft labels. As these soft labels are quantized for efficient storage~\cite{shen2022fast}, they must be recovered to their full dimension prior to calibration and averaging for final supervision. Specifically, we define the ensemble soft labels $ y_\text{\em en} = \frac{1}{M}\sum_{t \in \text{teachers}} {\hat y^t_a(x)}$, where $\hat y^t_a$ represents the recovered soft label with calibration from teacher $t$, $x$ is the input and $M$ is the number of teachers.

\begin{table}[t]
\centering
\vspace{-0.05in}
\resizebox{0.45\textwidth}{!}{
\begin{tabular}{c|c|c|c|c}
Mixup & CutMix & SelfMix (Ours) & ResNet50 & ViT-S/16 \\ \midrule[1pt]
 \checkmark  &            &              & 78.94 & 80.32  \\ 
             & \checkmark &              & 79.52 &  \bf 80.95 \\ 
             &            &  \checkmark  & \bf 80.47 & 79.29  \\ \midrule[0.5pt]
 \checkmark  &\checkmark  &              & 79.92 & \bf 81.16  \\ 
 \checkmark  &\checkmark  &  \checkmark  & \bf  80.26 & 80.69  \\ 
\end{tabular}
}
\vspace{-0.1in}
\caption{{Top-1 accuracy of data augmentations on different backbones}. On ViT, the soft label can benefit from combinations of more mixture operations. The backbones are ResNet-50 and ViT-S/16 and the teacher model is EfficientNet\_L2\_475. We run three trials and report the means.}
\label{tab:selfmix_result}
\vspace{-0.15in}
\end{table}

\section{ImageNet Experiments}

We conduct training on the ImageNet-1K (IN1K)~\cite{deng2009imagenet} training set. 
We report top-1 validation accuracy of a single 224$\times$224 crop. The default training budget is 300 epochs, and the temperature for both teacher and student is 1.0. The finetuning distillation settings follow their individual designs. All our soft label generation and model training are performed on the A100-GPU High-Flyer cluster with 80GB on each. More details are provided in Appendix.

\noindent{\bf Baselines: FKD}~\cite{shen2022fast} and {\bf FunMatch}~\cite{beyer2021knowledge}. Regions are randomly sampled in these two approaches. We use ResNet-50 and ViT-S/16 as the backbones for the ablation study. {\bf FKD + Curriculum Sampling}: (i) we sample regions for training from easier ones and gradually increase the level of difficulty. (ii) In contrast, we sample regions from hard ones and gradually decrease the level of difficulty.

\begin{table}[t]
\centering
\resizebox{0.49\textwidth}{!}{
\begin{tabular}{l|c|cc}
Teacher  &  $\text{T}_\text{Top-1}$ (\%)  &  $\text{S}_\text{ResNet50}$ (\%)  &  $\text{S}_\text{ViT-S/16}$ (\%) \\ \midrule[1pt]
Effi\_L2\_475~\cite{xie2020self}  &  88.14 & 80.23   & 81.16 \\  
Effi\_L2\_800~\cite{xie2020self}  &   \bf 88.39 &  80.16 & 81.30  \\ 
RegY\_128GF\_384~\cite{singh2022revisiting}  &  88.24  & \bf 80.34  & \bf 81.42 \\ \hline
ViT\_L16\_512~\cite{singh2022revisiting}  &   88.07 & \bf  80.29 & \bf 81.43 \\ 
ViT\_H14\_518~\cite{singh2022revisiting}  &    88.55 &  80.18  &  81.41 \\ 
BEIT\_L\_224~\cite{bao2021beit} &  87.52  & 80.03 & 81.16 \\  
BEIT\_L\_384~\cite{bao2021beit} &  88.40 &  80.06 & 81.11 \\  
BEIT\_L\_512~\cite{bao2021beit} &   88.60 & 80.09 & 81.07 \\  
ViT\_G14\_336\_30M~\cite{fang2022eva}  & 89.59  & 79.03 & 79.62 \\
ViT\_G14\_336\_CLIP~\cite{fang2022eva}  & 89.38  & 79.59 & 79.48 \\
\end{tabular}
}
\vspace{-0.1in}
\caption{Top-1 accuracy of distillation on ImageNet-1K using a {\bf singe} {\em ConvNet} or {\em Vision Transformer} teacher. Note that Mixup~\cite{zhang2018mixup} and CutMix~\cite{yun2019cutmix} are used for training ViT-S/16. We run three trials and report the means.}
\label{tab:single_teacher}
\vspace{-0.1in}
\end{table}

\subsection{Soft Label from Different Teachers}
The soft labels produced by distinct teacher models for the same input image may vary owing to differences in their distinctive features, architectures, and training strategies. In this section, our objective is to determine which teacher model has the greatest capacity to distill a student. We evaluate two types of student models: ResNet-50 and ViT-S/16. The results are shown in Table~\ref{tab:single_teacher}, where RegY\_128GF\_384 and ViT\_L16\_512 achieve the highest student accuracy individually, despite not being the best on their own.

\begin{table}[t]
\centering
\resizebox{0.486\textwidth}{!}{
\begin{tabular}{c|ccccc}
calibr. range   & [0, 0.2] & [0, 0.3] & [0, 0.4] & [0, 0.5] & [0,0.6] \\ \midrule[1pt]
    Top-1      &  80.31    & \bf  80.42  & 80.14 & 79.84 & 79.66   \\ 
\end{tabular}
}
\vspace{-0.1in}
\caption{Ablation of calibration for different probability ranges. The base model is the single teacher \texttt{FerKD} using ResNet-50 without \texttt{SelfMix} data augmentation.}
\label{tab:my_cal}
\end{table}

\begin{table}[t]
\centering
\vspace{-0.15in}
\resizebox{.37\textwidth}{!}{%
\begin{tabular}{c|c}
Pre-train     & Top-1 \\ \midrule[1pt]
  vanilla    &  	80.23    \\ 
+calibration\&selfmix     &   \ \ \ \ \ \ \ \ \ 80.68\textcolor{YellowGreen}{$^{\bf{+0.45}}$}  \\ 
+multi-teacher ensemble     & \ \ \ \ \ \ \ \ \  81.15\textcolor{YellowGreen}{$^{\bf{+0.47}}$}   \\ 
+more epochs       & \ \ \ \ \ \ \ \ \ 81.44\textcolor{YellowGreen}{$^{\bf{+0.29}}$}  \\ 
\end{tabular}
}
\vspace{-0.1in}
\caption{Ablation results using ResNet-50 on ImageNet-1K.}
\label{tab:my_tab_training_budget}
\vspace{-0.15in}
\end{table}

\begin{table*}[t]
\centering
\resizebox{0.98\textwidth}{!}{
\begin{tabular}{ccc|ccccc|c}
 Effi\_L2\_475        & Effi\_L2\_800 & RegY\_128GF\_384 & ViT\_L16\_512 & ViT\_H14\_518 & BEIT\_224 & BEIT\_384 & BEIT\_512 & Student Acc. \\ 
    \checkmark    &\checkmark &          &               &               &           &           &           &   80.23     \\ 
    \checkmark    &           &\checkmark&               &               &           &           &           &   \bf 80.52     \\ 
   \checkmark     &\checkmark & \checkmark&               &               &           &           &           &   80.49     \\ \hline
                  &           &           &               &\checkmark     &           &           &\checkmark &   80.29   \\ 
                  &           &           &  \checkmark   &\checkmark     &           &           & \checkmark&   \bf 80.53   \\ 
                  &           &           &  \checkmark   &\checkmark     & \checkmark& \checkmark&\checkmark &   80.48   \\  \hline
  \checkmark      &           &           &  \checkmark  &               &           &           &           &   80.35   \\ 
   \checkmark     &           &           &               &\checkmark     &           &           &           &   80.51     \\ 
   \checkmark     &           &           &               &               &           &           & \checkmark&   80.35     \\
                  &\checkmark &           &               &               &           &           & \checkmark&   80.38     \\ 
                  & \checkmark&           &               &\checkmark     &           &           &           &   80.51     \\ 
                  &           & \checkmark&  \checkmark   &               &           &           &           &   80.53     \\ 
                  &           & \checkmark&  \checkmark   & \checkmark    &           &           &           &   80.52  \\ 
  \checkmark      &           & \checkmark &  \checkmark  &               &           &           &           &   80.62   \\ 
  \checkmark      &           & \checkmark &  \checkmark  & \checkmark    &           &           &           &   \bf 80.74         \\                  
\end{tabular}
}
\vspace{-0.1in}
\caption{Ablation top-1 accuracy of teacher ensemble on ImageNet-1K with {\em ConvNet}, {\em Vision Transformer} or {\em hybrid teachers}. The left group is ConvNet teachers, the middle is the ViT teachers and the right group is the corresponding student ResNet-50 accuracy. Note that surgical calibration and \texttt{SelfMix} are not used here.}
\label{tab:ensemble_res}
\vspace{-0.22in}
\end{table*}

\subsection{Ablations}}
\noindent{\bf Ablation on Calibration.}
The results for different calibration ranges are shown in Table~\ref{tab:my_cal}, it can be observed that [0, 0.3] achieves the best accuracy. In practice, we discard examples in [0, 0.15) and (0.95, 1.0], meanwhile, calibrate examples in [0.15, 0.3] for the final strategy. More visualization of the hardest and easiest regions is shown in Fig.~\ref{fig:more_vis}.

\noindent{\textbf{Ablation on teacher ensemble and training budget.}} The results are in Table~\ref{tab:my_tab_training_budget}, showing that each design has the consistent improvement. Our final results are shown in Table~\ref{tab:final_res}, \texttt{FerKD} performs the best over other SOTA methods.

\vspace{0.1in}
\subsection{Curriculum Distillation}
 In curriculum distillation, the student model learns from the teachers' knowledge in a sequential manner, starting from easier examples (high probability regions) and gradually increasing the level of difficulty. This allows the student model to learn from easier to more complex regions. By presenting samples in a curriculum, the student model can learn from the easier samples and build a strong foundation before being exposed to more challenging samples. However, our experimental results, as shown in Table~\ref{tab:curriculum_dis}, demonstrate that this curriculum learning approach is inferior to our surgical label adaptation strategy, \texttt{FerKD}. Our approach outperforms curriculum distillation by 0.8\%. We attribute this performance improvement to the strong ability of soft labels to mitigate overfitting and improve generalization in the initial stages of training. As a result, the advantage of the curriculum learning approach is not as apparent in this scenario. We also notice that ``\texttt{e}-to-\texttt{h}'' performs slightly better than ``\texttt{h}-to-\texttt{e}'' demonstrating the effectiveness of curriculum strategy, while both of them are inferior to the random sampling baseline.

\begin{table}[t]
\centering
\resizebox{0.38\textwidth}{!}{
\begin{tabular}{l|c|c}
Method   & sampling  & Top-1 \\ \midrule[1pt]
Random (FKD~\cite{shen2022fast})    &  random    &  80.2   \\  
curriculum distillation       &   \texttt{h}-to-\texttt{e}  & 79.7  \\ 
curriculum distillation       &   \texttt{e}-to-\texttt{h}  & 79.9  \\
\rowcolor{mygray} \texttt{FerKD} (Ours)   &   surgical  &  \bf 80.7  \\ 
\end{tabular}
}
\vspace{-0.08in}
\caption{Curriculum and surgical (no ensemble) distillation. ``\texttt{e}-to-\texttt{h}'' refers to curriculum sampling from easy to hard regions. ``\texttt{h}-to-\texttt{e}'' refers to sampling from hard to easy.}
\label{tab:curriculum_dis}
\vspace{-0.1in}
\end{table}

\begin{table}[]
\centering
\resizebox{.48\textwidth}{!}{%
\begin{tabular}{l|c|c|c}
Model  & Epoch & Label    & Top-1 \\ \midrule[1pt]
Timm~\cite{wightman2021resnet}   & 600 & Hard$_\text{LS}$  &  80.4    \\ 
Pytorch (advanced)~\cite{resnetv2}   & 600 & Hard$_\text{LS}$  &   80.9   \\ \hline
MEAL~\cite{shen2019meal}   & \ 100$^\ddag$ & Soft  &    78.2  \\ 
MEAL V2~\cite{shen2020meal}  & \ 180$^\ddag$ & Soft  &  80.7    \\ 
MEAL V2~\cite{shen2020meal}$_\text{w/ CutMix}$  & \ 180$^\ddag$ & Soft  &   81.0    \\ \hline
ReLabel~\cite{yun2021re}$_\text{w/ CutMix}$  &  300  & Soft   &   80.2   \\ 
FunMatch~\cite{beyer2021knowledge}$_\text{w/ Mixup}$  &  300  & Soft   &   80.5   \\ 
FKD~\cite{shen2022fast}  &  300  & Soft   &   80.5   \\ 
\rowcolor{mygray} \texttt{FerKD} (Ours)   & 300  &  Adap &    \ \ \ \ \ \ \ 81.2\textcolor{YellowGreen}{$^{\bf{+0.7}}$}  \\  \hline
\rowcolor{mygray2}\texttt{FerKD} (Ours)   & 600  &  Adap &    \ \ \ \ \ \ \ 81.4\textcolor{YellowGreen}{$^{\bf{+0.9}}$}  \\
\end{tabular}
}
\vspace{-0.1in}
\caption{Top-1 accuracy on ImageNet-1K dataset. The backbone network in this table is ResNet-50. $\ddag$ indicates that the model was fine-tuned from hard-label pre-trained weights, resulting in a total training epoch of ``300 + $\times$''.}
\label{tab:final_res}
\vspace{-0.15in}
\end{table}

\subsection{Finetuning Distillation}

Finetuning distillation~\cite{shen2020meal} has been demonstrated as an effective approach to improve the accuracy of knowledge distillation (KD) frameworks. In the case of hard labels, multiple finetuning schemes have been proposed to adapt the model parameters to fit the target dataset, including partial finetuning on selected intermediate layers with varying learning rates~\cite{shen2021partial,aghajanyan2020better,lee2019would,ro2021autolr} or on the last few layers~\cite{yosinski2014transferable}. In this work, we adopt the MEAL V2~\cite{shen2020meal} protocol by finetuning the entire network from the pre-trained weights to evaluate the effectiveness of distillation with surgical soft label calibration as the objective. Specifically, we employ three types of pre-trained models:

{\bf (1)} Supervised pre-train: ResNet-50 on Timm~\cite{wightman2021resnet};

{\bf (2)} Weakly-supervised pre-train: RegNetY-128GF from SWAG~\cite{singh2022revisiting};

{\bf (3)} Self-supervised pre-train: ViT-G14 from EVA~\cite{fang2022eva};

We verify whether \texttt{FerKD} can continue improving fine-tuning distillation. 
As shown in Table~\ref{tab:finetuning_dis}, \texttt{FerKD} achieves consistent improvement across different architectures.

\begin{table}[]
\centering
\resizebox{.32\textwidth}{!}{
\begin{tabular}{c|c|c}
Pre-train & FT    & Top-1 \\ \midrule[1pt]
Timm~\cite{wightman2021resnet}      & Hard$_\text{LS}$  &  	80.38    \\ 
Timm~\cite{wightman2021resnet}      & FKD~\cite{shen2022fast}*   &  	80.62    \\ 
\rowcolor{mygray}Timm~\cite{wightman2021resnet}      & \texttt{FerKD}* (Ours) &    	\bf 81.06  \\ \hline
SWAG~\cite{singh2022revisiting}      & Hard$_\text{LS}$  &  	87.22    \\ 
SWAG~\cite{singh2022revisiting}      & FKD~\cite{shen2022fast}$^\dag$   &  	87.42    \\ 
\rowcolor{mygray}SWAG~\cite{singh2022revisiting}      & \texttt{FerKD}$^\dag$ (Ours) &    	\bf 87.76  \\ \hline
EVA\_MIM~\cite{fang2022eva}       &    Hard$_\text{LS}$   &   89.59   \\ 
EVA\_MIM~\cite{fang2022eva}       &   FKD~\cite{shen2022fast}    &   89.67   \\ 
\rowcolor{mygray}EVA\_MIM~\cite{fang2022eva}       &    \texttt{FerKD} (Ours)   &  \bf 89.86   \\ 
\end{tabular}
}
\vspace{-0.06in}
\caption{Fine-tuning distillation results using pre-trained ResNet-50 (Timm), RegNetY-128GF (SWAG) and ViT-G14 (EVA) on ImageNet-1K. $^\dag$ We use the same recipe of EVA for SWAG finetuning since SWAG did not provide the complete fine-tuning details. * On ResNet-50, we finetune with 150 epochs for both FKD and \texttt{FerKD}.}
\label{tab:finetuning_dis}
\vspace{-0.1in}
\end{table}

\subsection{Single Teacher {\em \bf vs.} Teacher Ensemble}

The soft labels generated by different teachers for the same input image can be different due to their unique features, architectures, and training strategies. By combining these soft labels from different teachers, the resulting label becomes more informative and can help the student model learn better representations. This approach can be particularly useful in scenarios where the teacher models have complementary strengths or when the input data is challenging and requires multiple perspectives to be accurately labeled. Furthermore, the use of multiple teachers can also help mitigate the effects of overfitting and improve the generalization performance of the student model. The results are shown in Table~\ref{tab:ensemble_res}, ensembling four hybrid teachers of {\em Effi\_L2\_475}, {\em RegY\_128GF\_384}, {\em ViT\_L16\_512}, and {\em ViT\_H14\_518} achieves the best accuracy.

\vspace{-0.08in}
\section{Transfer Learning Experiments}
\vspace{-0.05in}

\noindent{\bf Object detection and segmentation.} We conducted evaluations to investigate whether the improvement achieved by \texttt{FerKD} on ImageNet-1K can be transferred to various downstream tasks. Specifically, Table~\ref{tab:detection} presents the results of object detection and segmentation on COCO dataset~\cite{lin2014microsoft} using models pre-trained on ImageNet-1K with \texttt{FerKD}. We utilize Mask RCNN~\cite{he2017mask} with FPN~\cite{lin2017feature} following FKD for the experiment. Our \texttt{FerKD} pre-trained weight consistently outperforms both the baselines ReLabel and FKD on the downstream tasks. 
\begin{table}[]
\centering
\resizebox{0.33\textwidth}{!}{
\begin{tabular}{c|c|cc}
Method & Label    & $\text{AP}^\text{box}$  & $\text{AP}^\text{mask}$ \\ \midrule[1pt]
ReLabel~\cite{yun2021re}    &  LM    &   39.1  & 35.2   \\ 
FKD~\cite{shen2022fast}       &   Soft  &   39.7  & 35.9  \\ 
\rowcolor{mygray} \texttt{FerKD} (Ours)        &   Adap  &  \bf 40.2 & \bf 36.3   \\ 
\end{tabular}
}
\vspace{-0.1in}
\caption{COCO object detection and segmentation using a Mask-RCNN with FPN baseline. ``LM'' represents {\em label map} during pretraining. The backbone is ResNet-50-300ep.}
\label{tab:detection}
\vspace{-0.1in}
\end{table}

\begin{table}[]
\centering
\resizebox{0.48\textwidth}{!}{
\begin{tabular}{c|ccc}
Method & iNat 2019$_{224}$    & iNat 2019$_{336}$  & Places365$_{224}$ \\ \midrule[1pt]
EVA\_MIM~\cite{fang2022eva}     &   79.9  & 86.6  &  61.0   \\ 
\rowcolor{mygray}\texttt{FerKD} (Ours)    & \ \ \ \ \ \ \ \bf 80.3\textcolor{YellowGreen}{$^{\bf{+0.4}}$}  & \ \ \ \ \ \ \ \bf 87.1\textcolor{YellowGreen}{$^{\bf{+0.5}}$} & \ \ \ \ \ \ \ \bf 61.4\textcolor{YellowGreen}{$^{\bf{+0.4}}$}    \\ 
\end{tabular}
}
\vspace{-0.1in}
\caption{Transfer learning accuracy on various classification datasets. The input sizes are 224$\times$224 and 336$\times$336.}
\label{tab:transfer_learning}
\vspace{-0.15in}
\end{table}

\begin{table}[t]
\centering
\huge
\resizebox{.485\textwidth}{!}{
\begin{tabular}{l|c|cccc}
\multirow{2}{*}{Method}  & \multirow{2}{*}{IN1K} & \multirow{2}{*}{ReaL} & ImageNetV2  & ImageNetV2 & ImageNetV2   \\ 
        &      &    & Top-images & Matched-freq & Threshold-0.7 \\ 
        \midrule[2pt]
\multicolumn{6}{c}{{ResNet-50}: }\\
ReLabel~\cite{yun2021re} &   78.9  & 85.0 &  80.5 & 67.3 &  76.0   \\ 
 FKD~\cite{shen2022fast}     &    80.1  &   85.8  &  81.2  &   68.2  &    76.9  \\ 
\rowcolor{mygray}\bf \texttt{FerKD}     &    81.2  &  \ \ \ \ \ \ \ \bf 86.4\textcolor{YellowGreen}{$^{\bf{+0.6}}$}   & \ \ \ \ \ \ \ \bf 82.1\textcolor{YellowGreen}{$^{\bf{+0.9}}$}   & \ \ \ \ \ \ \ \bf  69.5\textcolor{YellowGreen}{$^{\bf{+1.3}}$}  &  \ \ \ \ \ \ \ \bf 77.8\textcolor{YellowGreen}{$^{\bf{+0.9}}$}    \\ 
\midrule[2pt] 
\multicolumn{6}{c}{ViT-G14-336:} \\
EVA~\cite{fang2022eva} &   89.6  &   90.8   &  89.0    &    81.9 &   86.7   \\ 
\rowcolor{mygray}\bf \texttt{FerKD}     &   89.9       &  \ \ \ \ \ \ \ \bf 91.3\textcolor{YellowGreen}{$^{\bf{+0.5}}$}     & \ \ \ \ \ \ \ \bf 89.4\textcolor{YellowGreen}{$^{\bf{+0.4}}$}    &  \ \ \ \ \ \ \ \bf 82.4\textcolor{YellowGreen}{$^{\bf{+0.5}}$}  &  \ \ \ \ \ \ \ \bf 87.1\textcolor{YellowGreen}{$^{\bf{+0.4}}$}  \\ 
\end{tabular}
}
\caption{Results of {\texttt{FerKD}} on ImageNet ReaL~\cite{beyer2020we} and ImageNetV2~\cite{recht2019imagenet} with ResNet-50 and ViT-G14 backbones.} 
\label{tab:my-table_INV2}
\vspace{-0.1in}
\end{table}

\noindent{\bf Classification tasks.} Table~\ref{tab:transfer_learning} shows the transfer learning result on  iNaturalists~\cite{ina2018} and Places~\cite{zhou2014learning} datasets. On both of these two datasets, our results surpass the baseline EVA pretrained model by significant margins.

\section{Robustness}

We provide comparisons on ImageNet ReaL~\cite{beyer2020we} and ImageNetV2~\cite{recht2019imagenet} datasets to examine the robustness of \texttt{FerKD} trained models. 
On ImageNetV2~\cite{recht2019imagenet}, we verify our \texttt{FerKD} models on three metrics ``Top-Images'', ``Matched Frequency'', and ``Threshold 0.7'' following FKD~\cite{shen2022fast}. 
We perform experiments on two network structures: ResNet-50 and ViT-G14. The results are shown in Table~\ref{tab:my-table_INV2}, we achieve consistent improvement over ReLabel and FKD on ResNet-50 (224$\times$224) and better accuracy than EVA on ViT-G14 (336$\times$336).

\section{Conclusion}

 In this work, we have presented a new paradigm of {\em faster knowledge distillation} ({\texttt{FerKD}}), which employs label adaptation on randomly cropped regions. The proposed method outperforms existing state-of-the-art distillation approaches in terms of both training speed and convergence. 
 Additionally, we make two key observations that could be leveraged in future studies. Firstly, we notice that the most challenging and easiest few crops obtained through the {\em RandomResizedCrop} operation do not contribute to the model's learning and can thus be discarded. Secondly, we find that moderate hardness crops can provide crucial context information that helps calibrate the model to learn more robust representations, which in turn benefit downstream tasks.

{\small
\bibliographystyle{ieee_fullname}
\bibliography{egbib}
}

\newpage
\appendix

\section*{Appendix}

In the appendix, we provide more details omitted in the main paper, including:

• Section~\ref{details}: Implementation details.

• Section~\ref{more_vis}: More visualization of identified crops.

\begin{table}[h]
\centering
    \centering
    \resizebox{.35\textwidth}{!}{%
    \begin{tabular}{lcc}
    \toprule[1.1pt]
    Backbone        &  ResNet-50    & ViT-S/16  \\
    Epoch           & 300     &  300    \\
    Batch size     & 1,024   & 1,024      \\
    Optimizer     & AdamW   &   AdamW   \\
    Init. {\em lr}   & 0.002  &   0.002   \\
    {\em lr} scheduler   & cosine   &  cosine    \\
    Weight decay   &  0.05   &  0.05     \\
    Warmup epochs   &  5   &  5   \\
    Num crops   &    4   &  4    \\ \hline
    Label smoothing   &  \xmark   &    \xmark    \\
    Dropout   &    \xmark   &  \xmark     \\
    Stoch. Depth   &  \xmark  &  0.1       \\
    Repeated Aug   &  \xmark  & \xmark      \\
    Gradient Clip.  &  \xmark  & \xmark      \\
    Rand Augment   &   \xmark  & \xmark     \\
    Mixup prob.   &  \xmark    &   0.8 \\
    Cutmix prob.   &  \xmark   &   1.0    \\
    SelfMix prob. & 1.0  &   \xmark\\
    Random erasing & \xmark  & \xmark \\
    \bottomrule[1.1pt] 
    \end{tabular}
    }
    \caption{Pre-training setting for ImageNet-1K.}
    \label{tab:my-table_distillation}
    \vspace{-0.1in}
\end{table}

\begin{table}[t]
\centering
\tablestyle{6pt}{1.15}
\scriptsize
\resizebox{.49\textwidth}{!}{%
\begin{tabular}{lc}
\toprule[0.9pt]
Backbone & ViT-G/14~\cite{fang2022eva} $|$ RegY-128GF~\cite{singh2022revisiting} \\
Peak learning rate & 3e-5 \\
Optimizer & AdamW \\
Optimizer hyper-parameters & $\beta_1$, $\beta_2$, $\epsilon$ = 0.9, 0.999, 1e-8 \\
Layer-wise lr decay & 0.95 \\
Learning rate schedule & cosine decay \\
Weight decay & 0.05 \\
\hline
Input resolution & 336 \\
Batch size & 512 \\
Warmup epochs & 2 \\
Training epochs & 15 \\
Num crops  &  2  \\
\hline
Drop path & 0.4 $|$ 0.0 \\
Augmentation & RandAug (9, 0.5) \\
Label smoothing & \xmark \\
Cutmix & \xmark \\
Mixup & \xmark \\
Random erasing & \xmark \\
SelfMix prob. & 1.0  \\
Random resized crop & (0.08, 1) \\
Ema & 0.9999 \\
Test crop ratio & 1.0 \\
\bottomrule[0.9pt] 
\end{tabular}
}
\vspace{-.5em}
\caption{Fine-tuning setting for ImageNet-1K.}
\label{1k_finetuning}
\vspace{-0.1in}
\end{table}

\begin{figure*}[t]
\begin{center}
\includegraphics[width=0.99\textwidth]{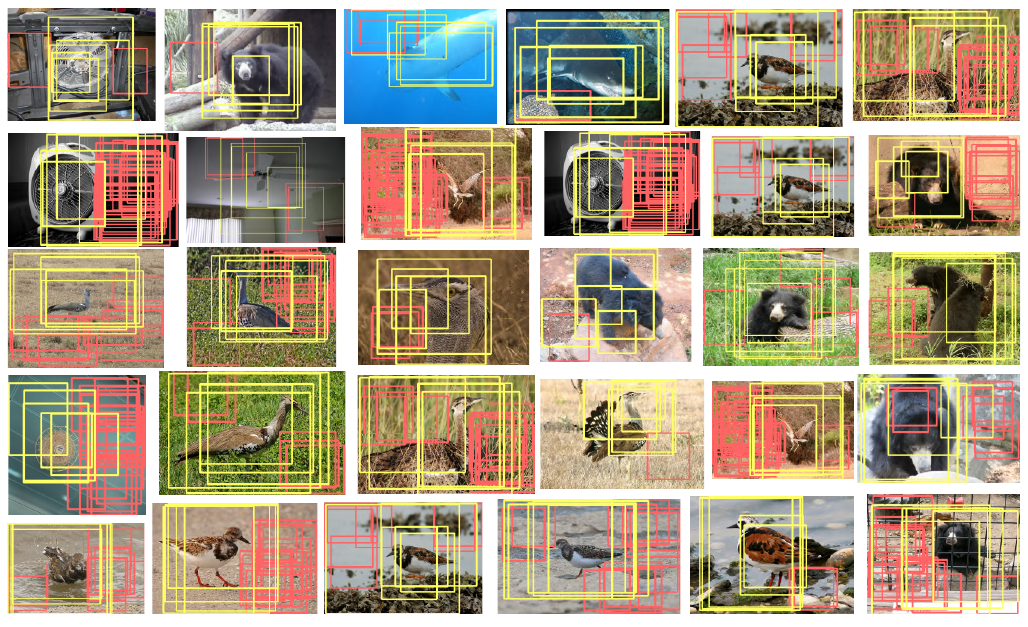}
\end{center}
\vspace{-0.23in}
   \caption{Illustration of the identified crops by teacher model for \textcolor{red}{hard} (background) and \textcolor{yellow}{easy} (foreground) samples. The teacher's probability can reflect object and background areas visually based on their magnitudes.}
\label{fig:more_vis_}
\vspace{-0.2in}
\end{figure*}

\section{Implementation Details} \label{details}

\noindent{\textbf{Training details for ResNet-50 and ViT-S/16 in the main text.}} We elaborate the detailed training settings and hyper-parameters of \texttt{FerKD} for pre-training from scratch on ImageNet-1K with ResNet-50 and ViT-S/16 backbones, as provided in Table~\ref{tab:my-table_distillation}. Generally, the training protocal follows FKD~\cite{shen2022fast}'s training strategy on ViT, DeiT and SReT. We employ \texttt{SelfMix} for ResNet-50, Mixup and CutMix for ViT-S/16 separately. We also use 4 as the number of crops in each image, batch size $= 1,024$ during training. 

\noindent{\textbf{Training details for finetuning ViT-G/14 and RegY-128GF in the main text.}} The finetuning settings and hyper-parameters of \texttt{FerKD} with ViT-G/14~\cite{fang2022eva} and RegY-128GF~\cite{singh2022revisiting} backbones are provided in Table~\ref{1k_finetuning}, which are similar to the training protocol in EVA~\cite{fang2022eva}. We employ \texttt{SelfMix} for both of the two pretrained backbones.

\noindent{\textbf{Data augmentation details for Mixup, Cutmix and \texttt{SelfMix}.}} The data augmentation configurations adopted in training are: for Mixup, we use probability 0.8 to generate the {\em Beta distribution}, and 1.0 for CutMix and \texttt{SelfMix}.

\section{More Visualization} \label{more_vis}

Fig.~\ref{fig:more_vis_} illustrates the identified crops by teacher for \textcolor{red}{hard} and \textcolor{yellow}{easy} samples. We do not involve any localization information, but the teacher's probability can reflect object and background areas automatically based on their magnitudes.

\end{document}